\documentclass[conference]{IEEEtran}

\usepackage[utf8]{inputenc}
\usepackage[T1]{fontenc}
\usepackage{cite}
\usepackage[nottoc,numbib]{tocbibind}
\usepackage[parfill]{parskip}
\usepackage{float}

\usepackage[hidelinks]{hyperref}
\usepackage{graphicx}

\usepackage{mathtools}
\usepackage[all]{nowidow}
\usepackage{xcolor}

\begin{document}

\hypersetup{pageanchor=false}

\title{Neural Network Exemplar Parallelization with Go}
\author{\IEEEauthorblockN{Georg Wiesinger}
\IEEEauthorblockA{\textit{Faculty of Computer Science} \\
\textit{University of Vienna}\\
Vienna, Austria\\
wiesingerg87@unet.univie.ac.at}
\and
\IEEEauthorblockN{Erich Schikuta}
\IEEEauthorblockA{\textit{Faculty of Computer Science} \\
\textit{University of Vienna}\\
Vienna, Austria \\
erich.schikuta@univie.ac.at}
}

\hypersetup{pageanchor=true}

\maketitle

\begin{abstract}
This paper presents a case for exemplar parallelism of neural networks using Go as parallelization framework. Further it is shown that also limited multi-core hardware systems are feasible for these parallelization tasks, as notebooks and single board computer systems.
The main question was how much speedup can be generated when using concurrent Go goroutines specifically. A simple concurrent feedforward network for MNIST\cite{MNISTData} digit recognition with the programming language Go\cite{Go,GoBlog,GoTour,GoGithubWiki} was created to find the answer.
The first findings when using a notebook (Lenovo Yoga 2) showed a speedup of 252\% when utilizing 4 goroutines. Testing a single board computer (Banana Pi M3) delivered more convincing results: 320\% with 4 goroutines, and 432\% with 8 goroutines.
\end{abstract}


\section{Introduction}
Neural networks and artificial intelligence are becoming more and more important not only in research, but also in daily used technology. Due to higher amounts of data these artificial agents have to analyze there is the need for a larger throughput and highly efficient neural networks. The programming language Go looks promising for developing such a highly efficient agent, as the language itself has been made not only for highly efficient parallelization but also with fast development in mind.
The main question is if Go is suitable for a highly efficient parallelization of neural networks.
The main objective is the creation of an efficient parallelized neural network. There is a possibility that Go could lead to a higher parallelization efficiency/speedup than other programming languages.
As Go is a young programming language
, the literature about this specific topic is very sparse to almost nonexistent. There are tertiary sources like websites comparing the general throughput of Go in comparison to web languages like NodeJS, PHP and Java\footnote{https://www.toptal.com/back-end/server-side-io-performance-node-php-java-go}. Other literature is related to parallelization speedup. There are also some neural networks realized in Go. No sources for a better comparison of parallelization techniques has been found.
The scope of this work is to find out the speedup when using multiple goroutines with a neural network while maintaining a high and sustainable classification accuracy. A working MNIST digit recognition system has been created for testing the speedup with up to sixteen cores. The network and parameters have been optimized, but due to only negligible improvements with more than 100 hidden layer nodes this amount has not been exceeded.
The execution time for one epoch has been sped up from 856.57-1271.73 (median 1005.80) seconds with 1 goroutine to only 171.50-221.38 (median 201.82) seconds with 16 goroutines with a Banana Pi M3. The Lenovo Yoga 2 showed a less significant speedup with 137.29-146.01 (median 142.33) for 1 goroutine to 55.10-64.62 (median 56.49) with 4 goroutines. Additional goroutines exceeding the maximum thread limit brought further speedup due to pipelining with the Banana Pi, but a negligible speed loss for the Lenovo Yoga.

\section{Related Work and Baseline Research}
Artificial neural networks and their parallel simulation gained high attention in the scientific community.
Parallelization is a classic approach for speeding up execution times and exploiting the full potential of modern processors. Still, not every algorithm can profit from parallelization, as the concurrent execution might add a non-negligible overhead. This can also be the case for data parallel neural networks, where accuracy problems usually occur, as the results have to be merged.

In the literature a huge number of papers on parallelizing neural networks can be found. An excellent source of references is the survey by Tal Ben-Nun and Torsten Hoefler~\cite{ben2019demystifying}.
However, only few research was done on using Golang in this endeavour.

In the following only specific references are listed, which influenced the presented approach directly.
The authors of \cite{liu2016accuracy} presented a parallel backpropagation algorithm dealing with the accuracy problem only by using a MapReduce and Cascading model.
In the course of our work on parallel and distributed systems~\cite{10.1007/BFb0057953,10.1007/3-540-62095-8_10,1380156} we developed several approaches for the parallelization of neural networks.
In \cite{schiki}, two novel parallel training approaches were presented for face recognizing backpropagation neural networks. The authors use the OpenMP environment for classic CPU multithreading and CUDA for parallelization on GPU architectures. Aside from that, they differentiated between topological data parallelism and structural data parallelism~\cite{schikuta1997structural}, where the latter is focus of the presented approach here. \cite{datavsmodel} gave a comparison of different parallelization approaches on a cluster computer. The results differed depending on the network size, data set sizes and number of processors. Using Go as parallelization tool was already analyzed for the Single-Program-Multiple-Data approach~\cite{kalwarowskyj2023spmd,turner2019go} and showed promising results. However, in this paper we focus on exemplar parallelization.
Besides parallelizing the backpropagation algorithm for training speed-up, alternative training algorithms like the Resilient Backpropagation described in \cite{rprop} might lead to faster convergence. One major difference to standard backpropagation is that every weight and bias has a different and variable learning rate. A detailed comparison of both network training algorithms was given in \cite{rpropcompare} in the case of spam classification.

\section{Data and Methods}
The following data and methods have been used to gain insight on the speedup possibilities.

\subsection{Choosing the data and parallelization method}
Different approaches of which data could be used have been evaluated. Weather data, crime rates, etc. all seemed to be a good fit, but with the possibility of very inconclusive outputs. Finally the "Hello World!" of neural networks has been chosen: The MNIST dataset\cite{MNISTData}. With this ready to use dataset the development process sped up as the convolutional part was already done.

Exemplar parallelism\cite{Rogers1997StrategiesFP} has been chosen as parallelization technique. Within the workers the learning method is stochastic gradient descent\cite{bottou2010large}, but due to combining the data in the main connectome, it behaves like a mini-batch update\cite{NIPS2010_4006}.

\subsection{Basic structure}
First a functional code for a basic neural network has been prepared. With this code it is also possible to define simple multi-layer feedforward networks. From that stable basis more functionality has been added (i.E. different activation functions) to ease up the future development.
Then the parallelization of the neural network has been implemented. There were additional challenges with avoiding possible race conditions. Go was very helpful with its built in race detector which can be utilized with the "-race" flag. It was easy to spot any race conditions and therefore the development sped up in the area deemed to take the most time.
Afterwards the possibility to input and compute large datasets has been implemented. A batch file functionality for ease of testing as well as data output functionality have been added too.
Afterwards the code and neural network have been optimized for a balance of speed, memory usage and training quality. Shuffling of the training data has been implemented to prevent any unwanted behavior that comes from repeated data. The elastic net regularization\cite{zou2005regularization} has been chosen to get better results and more stability for the neural network.

\subsection{The math}
Different activation functions have been tested to get a high accuracy, although this is not the purpose of this work. At the end it has been concluded that the best way is to start without any data normalization for the data to be put into the input layer. But before any activation function runs over the layer, the data is normalized by dividing each value by the size of the layer (including the bias) to minimize the risk of exploding gradients\cite{DBLP:journals/corr/IoffeS15}.

The following activation functions have been used:
\begin{itemize}
\item Input layer: Identity
\item Hidden layer: ELU\cite{DBLP:journals/corr/ClevertUH15}
\item Output layer: SoftMax
\end{itemize}

Variables are aas followed:
\begin{itemize}
\item $\eta$ = learning rate
\item t = target
\item x = neuron value before activation
\item $\varphi$ = activation function
\item $\delta$ = error
\item w = weight
\end{itemize}

\subsubsection{Activation functions}

\textbf{Identity}\\
The simplest activation function is the identity function. It simply states that the value stays the same, as in equation \eqref{Identity}.
\begin{equation}
\label{Identity}
\varphi(x) = x
\end{equation}

So the derivation simply is 1 as shown in equation \eqref{IdentityDerived}.
\begin{equation}
\label{IdentityDerived}
\varphi'(x) = 1
\end{equation}

\textbf{Exponential Linear Unit}\\
In comparison to ReLU and leaky ReLU, ELU "[...] speeds up learning in deep neural networks and leads to higher classification accuracies."\cite{DBLP:journals/corr/ClevertUH15} and therefore has been chosen over the other options. Equation \eqref{ELU} shows the math, where alpha is a positive value that can be freely chosen. 

\begin{equation}
\label{ELU}
\varphi(x) =\begin{cases}
x & \text{if x $\geq$ 0}\\
\alpha * (e\textsuperscript{x} - 1) & \text{if x < 0}
\end{cases}
\end{equation}

The derivation for training is shown in equation \eqref{ELUDerived}.

\begin{equation}
\label{ELUDerived}
\varphi'(x) =\begin{cases}
1 & \text{if x $\geq$ 0}\\
\varphi(x) + \alpha & \text{if x < 0}
\end{cases}
\end{equation}

\textbf{SoftMax}\\
The SoftMax function gives us a classification of the likelihood that the current input represents a certain number. The math is straightforward and shown in equation \eqref{SoftMaxOriginal}.

\begin{equation}
\label{SoftMaxOriginal}
\varphi\textsubscript{i}(\overrightarrow{x}) =
\frac{e\textsuperscript{x\textsubscript{i}}}{\sum\limits_{j=1}^J e\textsuperscript{x\textsubscript{j}}}
\end{equation}

But with this equation, exploding or vanishing gradients\cite{DBLP:journals/corr/abs-1211-5063} can become a problem due to the high likelihood of numbers exceeding.
For the SoftMax activation there is a little "trick". It is possible to add a scalar, as shown in \eqref{SoftMaxConstant}, without changing the value of the softmax function\cite{GoodfellowEtAl2016}.

\begin{equation}
\label{SoftMaxConstant}
\varphi\textsubscript{i}(\overrightarrow{x}) =
\frac{e\textsuperscript{x\textsubscript{i} + S}}{\sum\limits_{j=1}^J e\textsuperscript{x\textsubscript{j} + S}}
\end{equation}

So, instead of using softmax(x), softmax(z) - with a scalar value of the negative x maximum - has been used, as in equation \eqref{zEquation}.

\begin{equation}
\label{zEquation}
z\textsubscript{i} = (x\textsubscript{i} - max\textsubscript{i}(x\textsubscript{i}))
\end{equation}

If we use the maximum, we push the calculation into the negative number spectrum. So, instead of having values ranging over ]-$\infty$ , $\infty$[, they've been shifted to ]0, 1]
, as in \eqref{SoftMax}.

\begin{equation}
\label{SoftMax}
\varphi\textsubscript{i}(\overrightarrow{x}) =
\frac{e\textsuperscript{z\textsubscript{i}}}{\sum\limits_{j=1}^J e\textsuperscript{z\textsubscript{i}}}
\end{equation}

Equation \eqref{SoftMaxDerivation} derivation for training is a little bit more complicated.

\begin{equation}
\label{SoftMaxDerivation}
\varphi'\textsubscript{i}(\overrightarrow{x}) = \frac{\partial\varphi\textsubscript{i}(\overrightarrow{x})}{\partial x\textsubscript{j}} =\begin{cases}
\varphi\textsubscript{i}(\overrightarrow{x}) * (1 - \varphi\textsubscript{j}(\overrightarrow{x})) & \text{i = j}\\
\varphi\textsubscript{i}(\overrightarrow{x}) * (0 - \varphi\textsubscript{j}(\overrightarrow{x})) & \text{i $\neq$ j}
\end{cases}
\end{equation}

Mathematicans use \eqref{SoftMaxDelta} to shorten the equation to \eqref{SoftMaxWithDelta}.

\begin{equation}
\label{SoftMaxDelta}
\delta\textsubscript{ij} = \begin{cases}
1 & \text{i = j}\\
0 & \text{i $\neq$ j}
\end{cases}
\end{equation}

\begin{equation}
\label{SoftMaxWithDelta}
\varphi'\textsubscript{i}(\overrightarrow{x}) = \frac{\partial\varphi\textsubscript{i}(\overrightarrow{x})}{\partial x\textsubscript{j}} = \varphi\textsubscript{i}(\overrightarrow{x}) * (\delta\textsubscript{ij} - \varphi\textsubscript{j}(\overrightarrow{x}))
\end{equation}

But in the end it comes down to the same function as the logistic derivation, shown in equation \eqref{SoftMaxDerived}. As the result of the derivation is a diagonal matrix\cite{NIPS1993_877} there is no need to calculate the whole matrix.

\begin{equation}
\label{SoftMaxDerived}
\varphi'\textsubscript{i}(\overrightarrow{x}) = (1 - \varphi\textsubscript{i}(\overrightarrow{x})) * \varphi\textsubscript{i}(\overrightarrow{x}) = (1 - x) * x
\end{equation}

\subsubsection{Elastic Net Regularization}
The elastic net regularization\cite{zou2005regularization} has been used for weight updates.

It's a combination of the lasso regression \eqref{LassoRegression} and the ridge regression \eqref{RidgeRegression}.

\begin{equation}
\label{LassoRegression}
L\textsuperscript{1} = \lambda|w|
\end{equation}

\begin{equation}
\label{RidgeRegression}
L\textsuperscript{2} = \lambda w\textsuperscript{2}
\end{equation}

The elastic net \eqref{ElasticNet} is simple.

\begin{equation}
\label{ElasticNet}
ElasticNet = \lambda|w| + \lambda w\textsuperscript{2}
\end{equation}

Computational optimization \eqref{ElasticNetOptimized}

\begin{equation}
\label{ElasticNetOptimized}
ElasticNet = \lambda (|w| + w\textsuperscript{2})
\end{equation}

For the derivation \eqref{SignumDerivation} the signum function \eqref{Signum} is needed.

\begin{equation}
\label{SignumDerivation}
|w| = w * sgn(w)
\end{equation}

\begin{equation}
\label{Signum}
sgn(w) = \begin{cases}
1 & \text{w > 0}\\
0 & \text{w = 0}\\
-1 & \text{w < 0}
\end{cases}
\end{equation}

Which leads to \eqref{ElasticNetDerived}.

\begin{equation}
\label{ElasticNetDerived}
ElasticNet' = \lambda(sgn(w) + 2w)
\end{equation}

\subsubsection{Loss function}
Quadratic loss has been chosen as loss function. Although the classification of handwritten digits - as the name says - is a classification problem and therefore cross entropy loss should show better results. Different loss functions will be implemented in the future of this work.

\begin{equation}
\label{LossFunction}
L = - \frac{1}{2} \sum\limits_{i}^{nclass} \Bigl(t\textsubscript{i} - \varphi(x\textsubscript{i})\Bigr)\textsuperscript{2} + \lambda \frac{1}{2} \sum\limits_{i}^{k} \Bigl( |w\textsubscript{i}| + w\textsubscript{i}\textsuperscript{2} \Bigr)
\end{equation}


The derivative is the logistic equation, therefore the loss function is equation \eqref{LossFunctionDerived}.

\begin{equation}
\label{LossFunctionDerived}
L' = - \sum\limits_{i}^{nclass} \Bigl(t\textsubscript{i} - \varphi(x\textsubscript{i})\Bigr) + \lambda \sum\limits_{i}^{k} \Bigl( \frac{1}{2}sgn(w\textsubscript{i}) + w\textsubscript{i}) \Bigr)
\end{equation}


\subsubsection{Forward and backward propagation}
All the previous information is needed to understand the forward and backward propagation methods.

\textbf{Forward}\\
After setting the inputs and targets, the first layer of the neural network gets activated. Then for each layer, the next neuron gets excited with the product of the activated value and the weight of the connection between the neurons \eqref{ExciteNeuron}.
\begin{equation}
\label{ExciteNeuron}
x\textsubscript{j} = \varphi(x\textsubscript{i})*w\textsubscript{ij}
\end{equation}

\textbf{Backward}\\
When learning, equation \eqref{Error} is used to calculate the error.

\begin{equation}
\label{Error}
\delta = t - \varphi(x)
\end{equation}

The formula for the weight update, with learning rate and the regularization in \eqref{weightDelta}.

\begin{equation}
\label{weightDelta}
\Delta w = \eta * (\delta * \varphi(x) + \lambda (sgn(w) + w))
\end{equation}

The final equation in \eqref{weightUpdateEquation}.

\begin{equation}
\label{weightUpdateEquation}
w\textsubscript{ij}\textsuperscript{+} = w\textsubscript{ij} - \Delta w,
\end{equation}

\subsection{Choosing the parameters}
ELU alpha = 0.5 (currently hardcoded)\\
The alpha value for ELU has been hardcoded as there was no incentive to do otherwise in the current software iteration.

\textbf{workerBatch:} 100\\
The worker batch has been chosen to merge the single neural networks as often as possible, but without losing too much performance due to context switching.

\textbf{Minimum/Maximum weight (starting weights):} [-0.1; 0.1]\\
As the weights are usually getting smaller, when learning occurs, the starting values have to be chosen to be 0.1 instead of 1. This led to the best outcome.

\textbf{LearningRate:} 0.8\\
The learning rate has been set to 0.8, as this led to the best outcome.

\textbf{Lambda:} 0.0000001\\
The multiplicator for the elastic net has been set to this value, as it provided the highest accuracy for the training and test set. As it is hard to tell if either L1 or L2 regularization is the best, there is only one lambda for setting both methods, to achieve a balance between the two methods.

\section{Results/Evaluation}
For testing the neural network, two available systems have been chosen: The Lenovo Yoga 2 laptop, as it is a dual core consumer product which utilizes threading and a turbo mode for higher workloads, with 64 Bit Linux. The Banana Pi M3, as it is a well known octa core home server, with no threading, and without data distortion due to turbo mode kicking in, and 32 Bit Linux. Both systems have a standard CPU frequency of 1.80 GHz, although the minimum and maximum values differ.

There are stark differences in computation speed as well as speedup between the Intel and the ARM architecture. As RISC and CISC lost their meaning to describe newer architectures, it is not possible to draw the conclusion here
, although the main effect could come from the smaller - and therefore faster access rates - of the Intel L1 and L2 caches, or the lack of an L3 cache in the ARM architecture. Further research would be needed.

\subsection{Benchmark}
\begin{figure}[H]
	\centering
	\includegraphics[width=1.0\columnwidth]{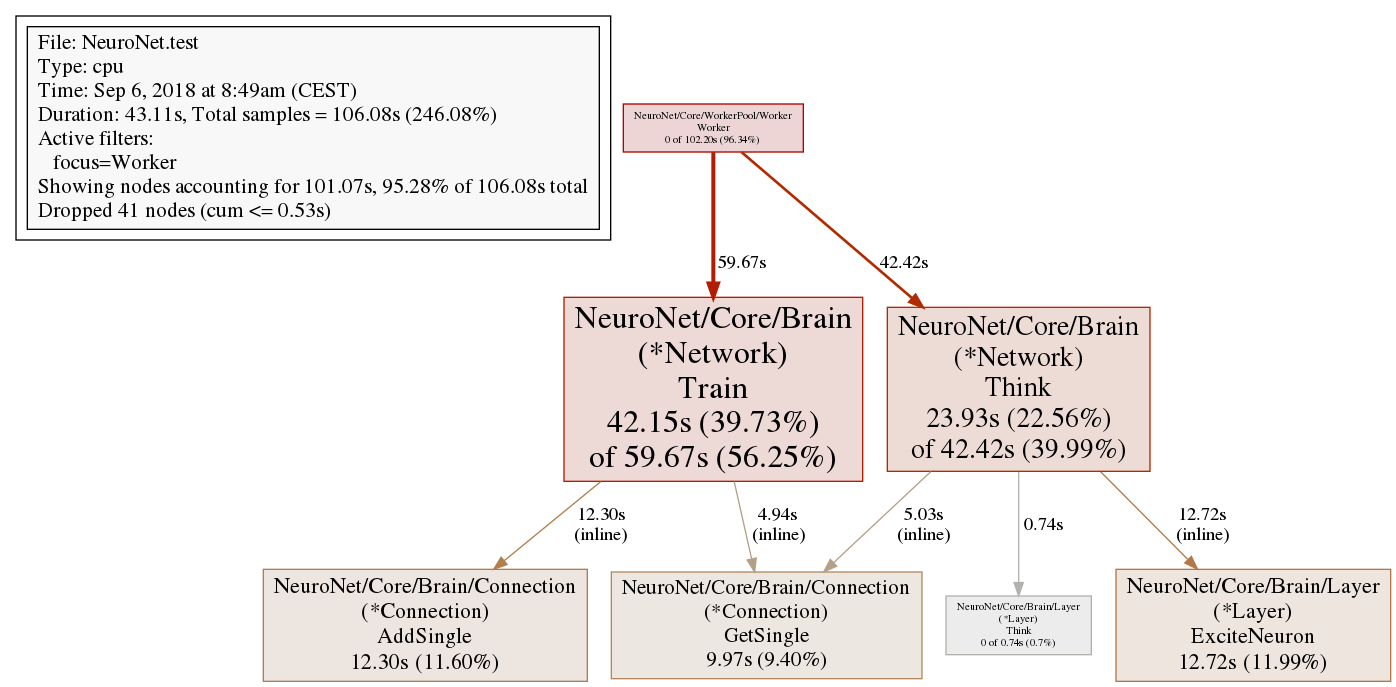}
	\vspace{-1em}
	\caption{pprof worker CPU profile}
	\vspace{-0.5em}
	\label{fig:pprof}
\end{figure}

When using pprof for checking the total cpu usage of the code parts with BenchBatch (it utilizes 4 cores, uses a worker batch of 100 lines, and processes 1 training with 60.000 MNIST lines as well as 1 test with 10.000 MNIST lines), it can be seen in \autoref{fig:pprof} that thinking and training takes up about 96\% of the total time. Thinking takes about 40\% of the time, training takes about 56\%. Thinking is the forward propagation, training is the backward propagation. Due to that high amount of cpu usage heavy optimizations were made in these code parts, as these had the greatest effects.

The utility functions only play a marginal role. Even though they wouldn't need any optimization, they've been optimized for general code quality reasons. In example the garbage collector (mallocgc) is hardly used in the utility functions, and almost never in the main code part. As strings are only converted when needed, these parts of the code - even though they're not impacting the measurements - have been highly optimized. Maybe there's still room for further optimization, but for the general purpose this goal has been exceeded.

\subsection{Test systems}
The final tests were made with a "Lenovo Yoga 2 Pro Multimode Ultrabook" as well as a "Banana Pi M3". 

Specifications of the Lenovo:
Intel(R) Core(TM) i7-4500U CPU @ 1.80GHz, Dual Core (4 threads)\\
Min CPU: 800 MHz, Max CPU: 3.0 GHz\\
32 KiB L1 cache, 256 KiB L2 cache, 4 MiB L3 cache\\
2x4096 DIMM @ Clockspeed 1600 MHz\\
64 Bit Linux Ubuntu 18.04.1 LTS

Specifications of the Banana Pi M3:\\
A83T ARM Cortex-A7 octa-core CPU @ 1.80 GHz, Octa Core (8 threads), 4800 BogoMIPS\\
ARMv7 Processor rev 5 (v7l)\\
Min CPU: 480 MHz, Max CPU: 1.8 GHz\\
512 KiB L1 cache, 1 MiB L2 cache\\
2GB LPDDR3\\
32 bit (armv7l) Linux Ubuntu 16.04.5 LTS, MATE Desktop Environment 1.12.1

\subsubsection{Lenovo Yoga 2}
\begin{figure}[htpb]
	\centering
	\includegraphics[width=1.0\columnwidth]{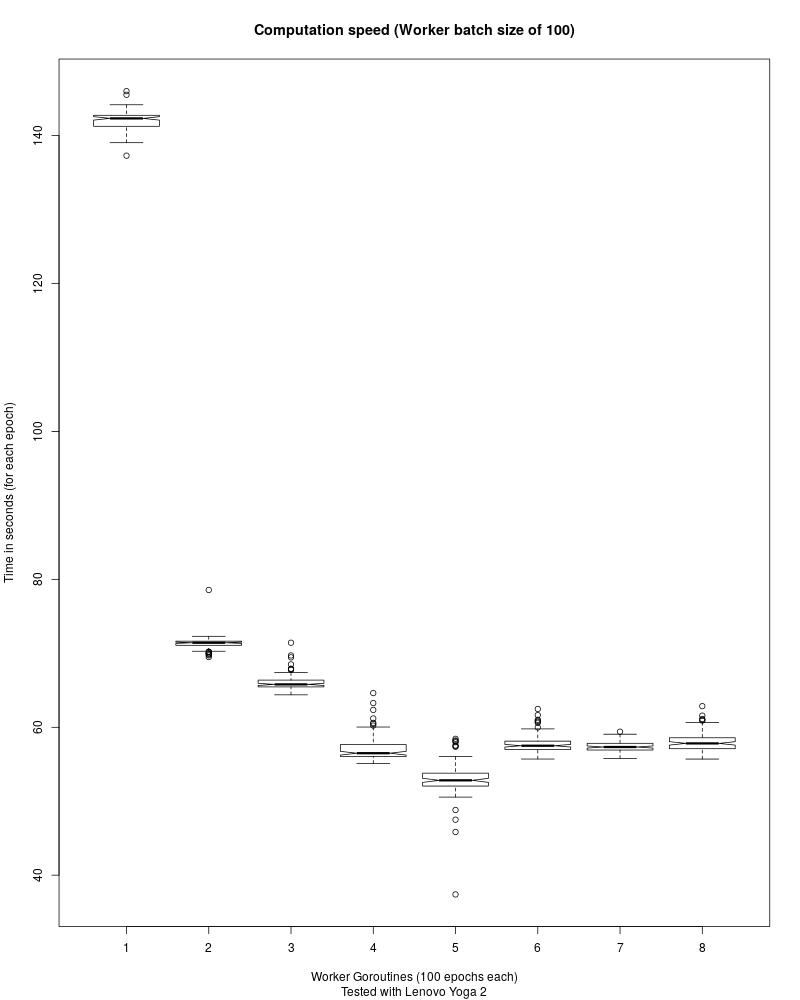}
	\vspace{-1em}
	\caption{Lenovo Yoga 2 speedup}
	\vspace{-0.5em}
	\label{fig:lenovoAllGoroutines}
\end{figure}

The 252\% speedup generated with 4 goroutines on the Lenovo Yoga 2 when utilizing more than 1 processor is clearly visible in \autoref{fig:lenovoAllGoroutines}.
It is also visible that using more goroutines than processors slows the execution time down only by an almost negligible amount.

\begin{figure}[htpb]
	\centering
	\includegraphics[width=1.0\columnwidth]{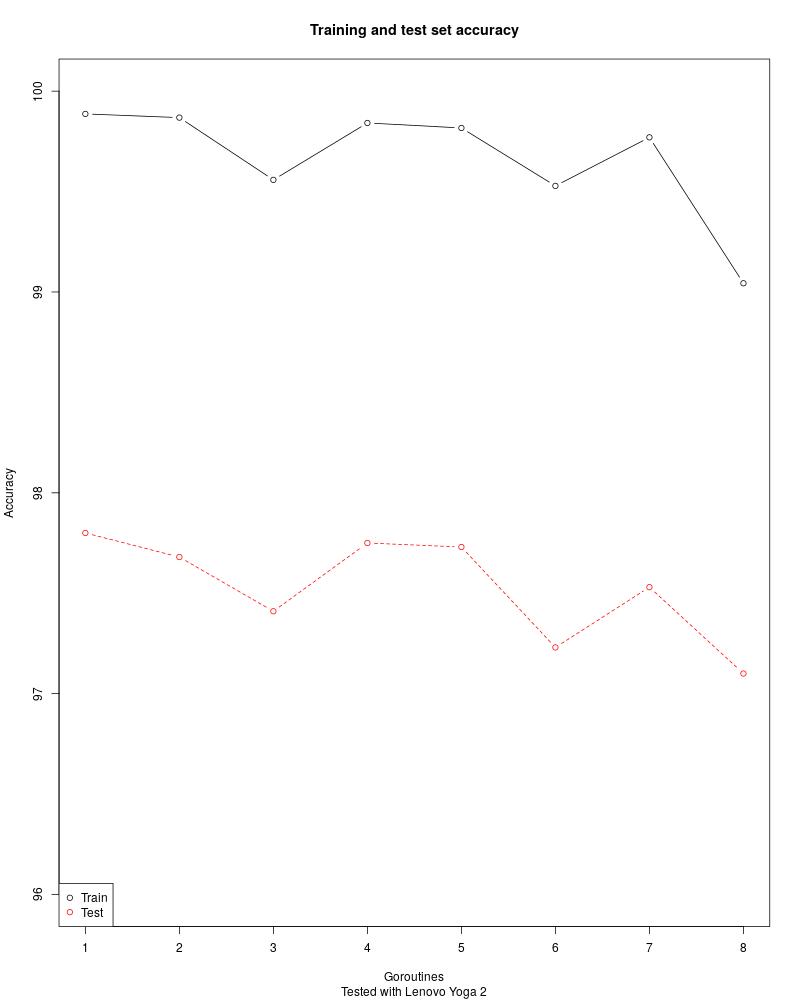}
	\vspace{-1em}
	\caption{Lenovo Yoga 2 accuracy}
	\vspace{-0.5em}
	\label{fig:lenovoMaxTestTrain}
\end{figure}

Parallelization speedup comes at a price. Although very small, there is a slight decrease in recognition rates when utilizing more goroutines as shown in \autoref{fig:lenovoMaxTestTrain}.

\subsubsection{Banana Pi M3}
\begin{figure}[htpb]
	\centering
	\includegraphics[width=1.0\columnwidth]{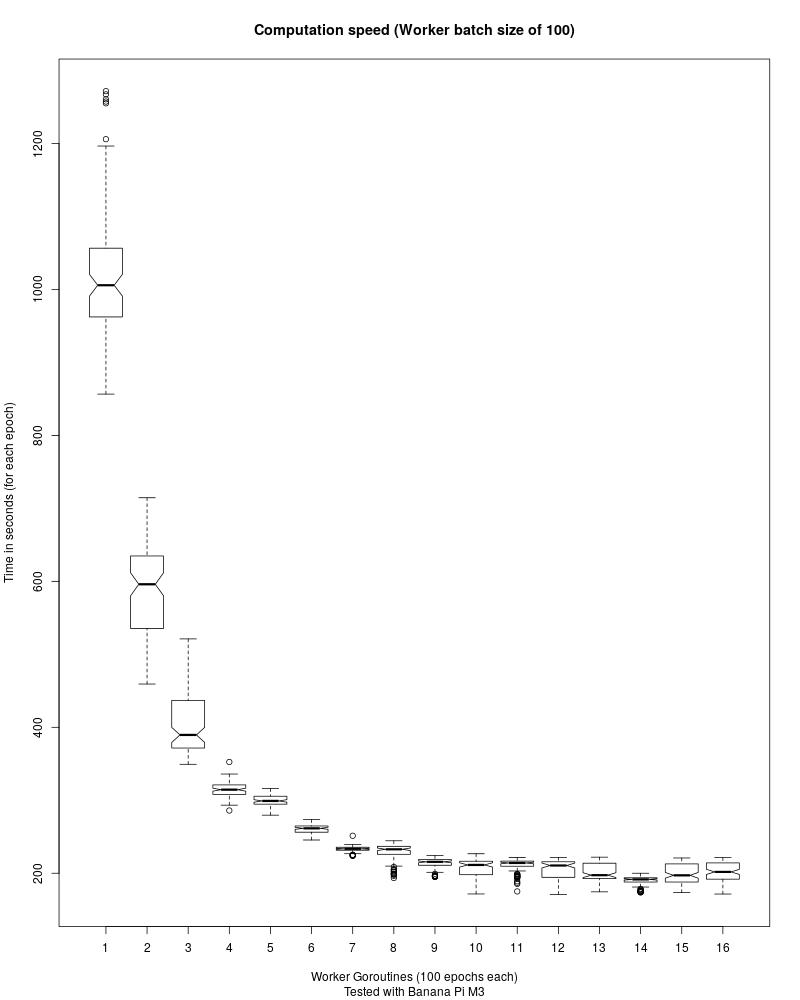}
	\vspace{-1em}
	\caption{Banana Pi M3 speedup}
	\vspace{-0.5em}
	\label{fig:bpiAllGoroutines}
\end{figure}

When looking at the results of the Banana Pi M3 in \autoref{fig:bpiAllGoroutines}, it is apparent that utilizing multiple cores leads to an even greater benefit than with the Lenovo. It was possible to generate a 320\% speedup with 4 goroutines, and - due to pipelining - it was even possible to generate over 498\% speedup when using more goroutines than there were threads available.

\begin{figure}[htpb]
	\centering
	\includegraphics[width=1.0\columnwidth]{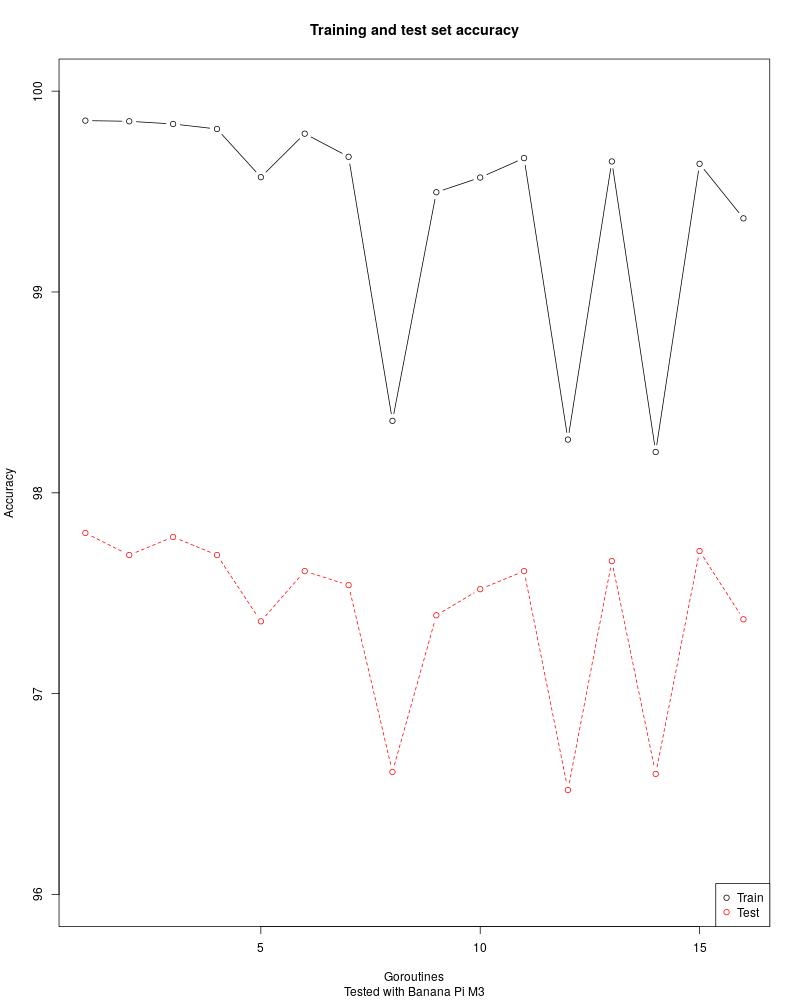}
	\vspace{-1em}
	\caption{Banana Pi M3 accuracy}
	\vspace{-0.5em}
	\label{fig:bpiMaxTestTrain}
\end{figure}

The training and test set accuracies look promising too. A 99.26\% training set accuracy and 97.14\% test set accuracy with only one core has been accomplished. The accuracy does not get lower when utilizing more cores, even though quality differences in the recognition rate can occur. In \autoref{fig:bpiMaxTestTrain} it is clearly visible that recognition rate drops can occur at any time.

\subsection{Accuracy growth depending on goroutines}
\begin{figure}[htpb]
	\centering
	\includegraphics[width=1.0\columnwidth]{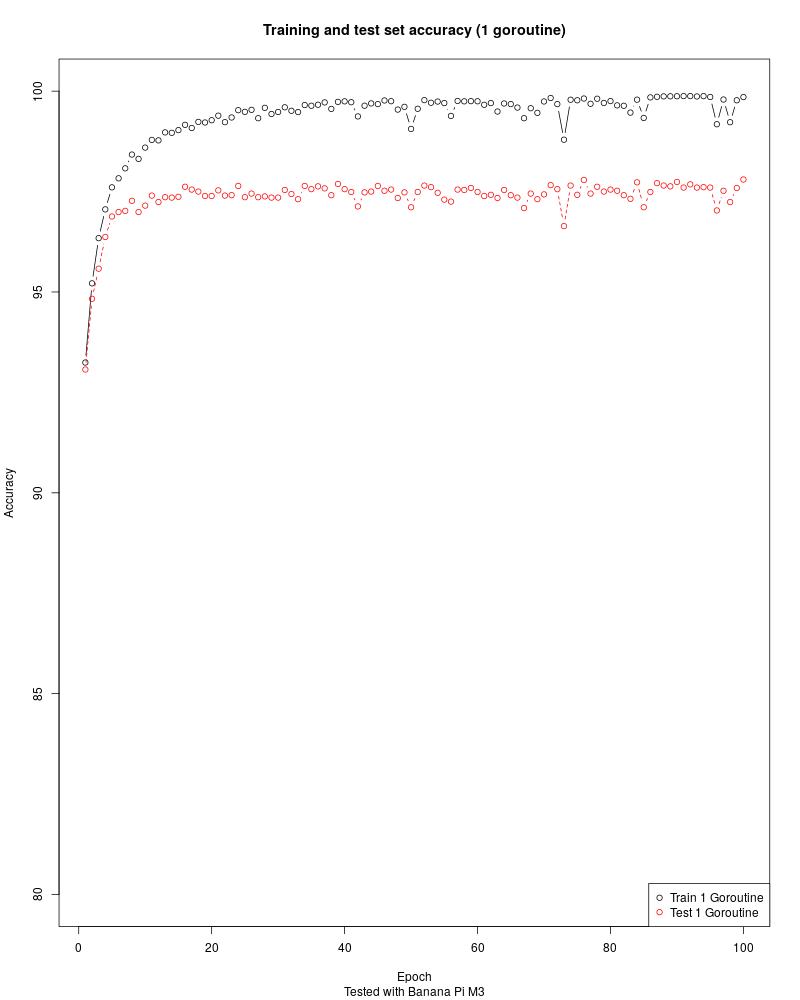}
	\vspace{-1em}
	\caption{Accuracy growth with 1 Goroutine}
	\vspace{-0.5em}
	\label{fig:bpiAccuracy1}
\end{figure}

When only one goroutine is used\autoref{fig:bpiAccuracy1} with the Banana Pi, the neural network starts with a very high recognition accuracy after the first epoch and has a very good learning rate.

\begin{figure}[htpb]
	\centering
	\includegraphics[width=1.0\columnwidth]{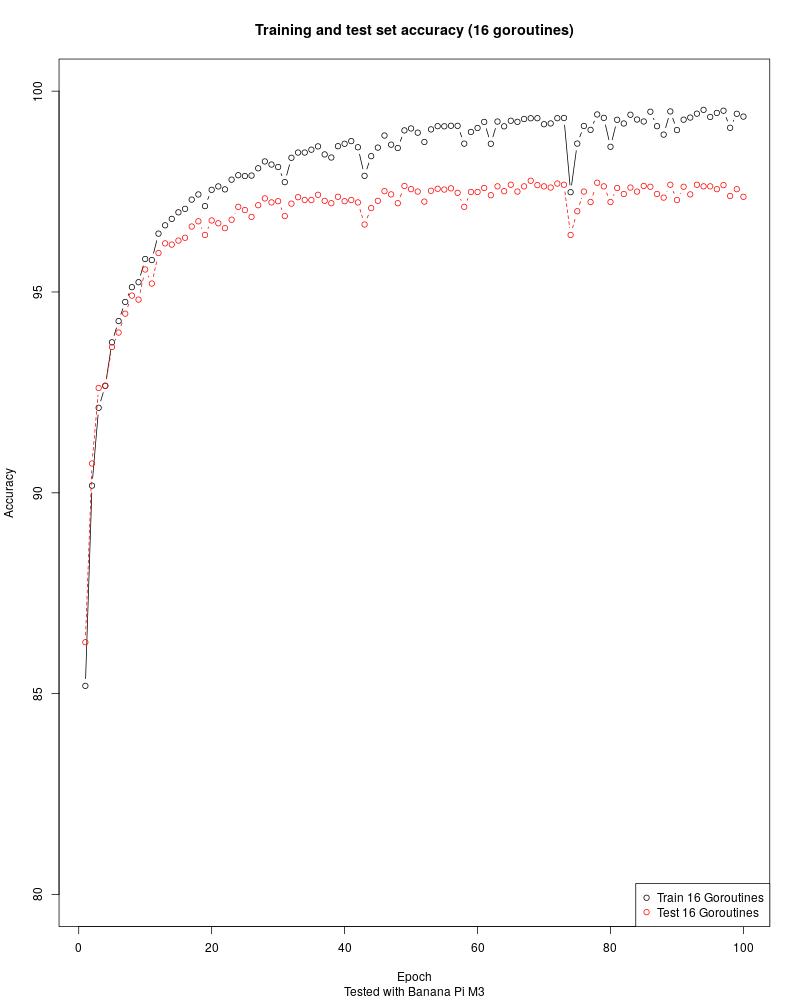}
	\vspace{-1em}
	\caption{Accuracy growth with 16 goroutines}
	\vspace{-0.5em}
	\label{fig:bpiAccuracy16}
\end{figure}

With 16 goroutines\autoref{fig:bpiAccuracy16} the recognition accuracy starts lower and the network takes longer to learn.

\textbf{Recognition rates in a nutshell}

1 Goroutine, accuracy > 90\%/95\%/99\%:\\
93.24\% accuracy after 1 epoch, 1040 seconds\\
95.22\% accuracy after 2 epochs, 2006 seconds\\
99.03\% accuracy after 15 epochs, 15608 seconds

16 goroutines, accuracy > 90\%/95\%/99\%:\\
90.18\% accuracy after 2 epochs, 392 seconds\\
95.12\% accuracy after 8 epochs, 1583 seconds\\
99.02\% accuracy after 49 epochs, 9628 seconds

To reach a higher accuracy with more goroutines more epochs and training samples are needed. But the speedup allows to train it in shorter time - or, to look at it from another perspective - to compute more inputs in a much shorter timespan.

\section{Lessons Learned}
The final part of this work is to look at what has been learned about the "do's and don'ts of implementing neural networks", Go as a language, the drawn conclusion, and possible future work.

\subsection{"To do, or not to do?" of implementing neural networks}
There are certain roads to victory and many paths to development hell. The latter leads to a steeper learning curve and should therefore be preferred when trying to understand the implications of certain design decisions - but under normal circumstances the beaten path is the quicker route. These recommendations for other coders shall make implementing neural networks a little bit easier and shine a light on which thought processes are good and which are impractical to do.

\subsubsection{Arrays instead of structs}
Do not use one struct instance per neuron, connection, etc. as it has a large overhead. The compiler is able to optimize the usage of arrays. The first iteration of the neural network took hours for just one epoch on the Lenovo, while the array version takes less than a minute.

\subsubsection{Only save when necessary}
Only save and load data when needed. In the context of the neural network: Save either batchwise or after every epoch. Try to hold the data in the memory as long as possible. 

\subsubsection{Machine readable is better than human readable}
The conversion of data to XML, JSON, or any other human readable format takes a higher amount of computation time, memory, and disk space, than machine readable formats.
If a human readable format is needed, it should only be created, if a human wants to read it and there is a need for them to do so. Sifting through millions of weights and updates is not something a human should do. But, depending on the use case, the human readable format can be created, when
\begin{itemize}
\item The process is finished and the results shall be shown.
\item An error occurs and the data is necessary to fix it.
\end{itemize}

If human entities want to access data while the process is running (in real time, or stepwise for debugging) there are different approaches:
\begin{itemize}
\item Create only one file every few epochs which can be accessed by multiple human entities. Do NOT create it for every entity that accesses the file.
\item Duplicate the machine readable results and parse them on a different system. For snapshots a simple ID can be given to every file.
\end{itemize}

\subsubsection{Parallelization and context switches}
It takes time to store states of threads. Data has to be shoved around between CPU caches. If applicable give a worker as much data as possible, with one drawback in mind: More data merges mean higher fluctuations and slower computation - less data merges can lead to a more stable convergence and faster computation\cite{DBLP:journals/corr/Ruder16}, as well as a higher level of generalization\cite{DBLP:journals/corr/IoffeS15}. All while being able to perform online learning due to the singular workers performing stochastic gradient descent.

\subsubsection{Struct packing}
In Go it is possible to pack structs. That means organizing the data types in a way so that they waste the least amount of memory. The principle for this work was "Memory is cheap, but even though students lack the money to buy some, there is no need to overdo it". But one late evening (at 7 o'clock in the morning) these principles had been thrown over board. So structs have been packed. 

\subsubsection{Slices}
\begin{figure}[htpb]
	\includegraphics[width=1.0\columnwidth]{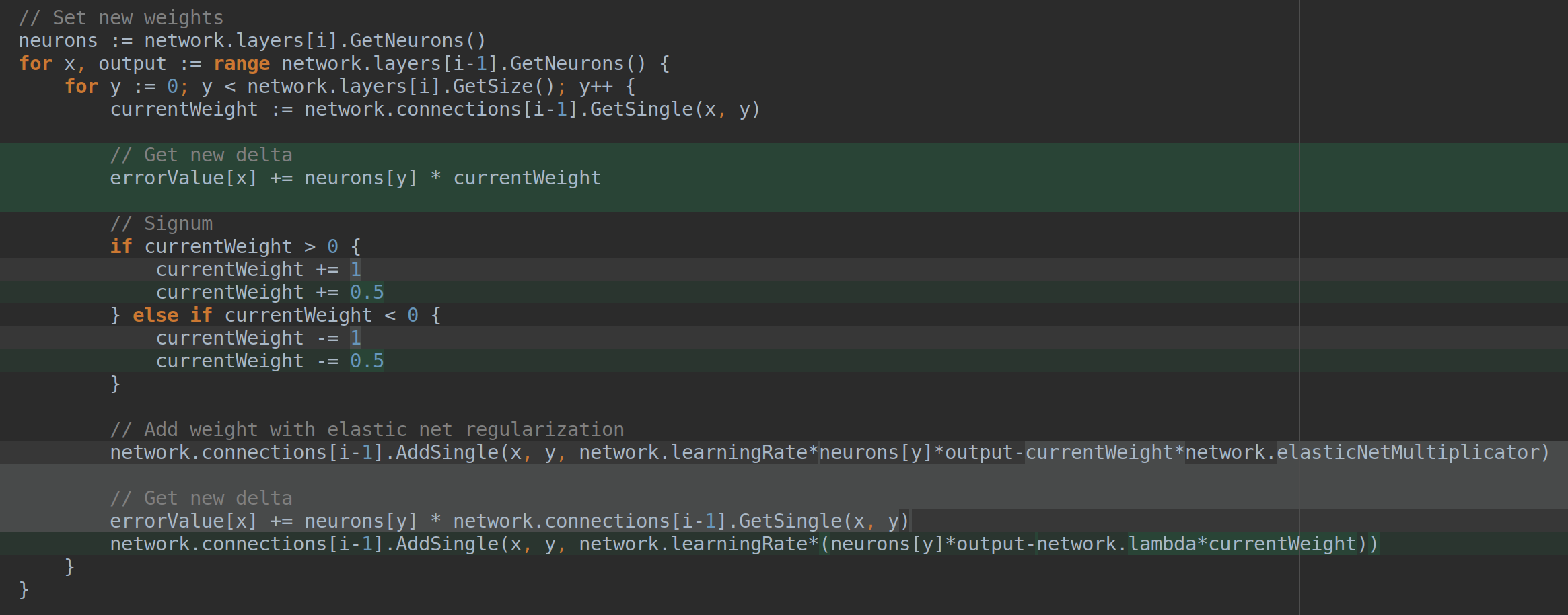}
	\vspace{-1em}
	\caption{Code changes from wrong to correct code}
	\vspace{-0.5em}
	\label{fig:changeToCorrect}
\end{figure}

Do not loop through lines of data to append it to a batch line by line. Use the slice functionality of Go - which passes them by reference - if applicable. The following data has been taken from the old model with using the updated weights for the error calculation \autoref{fig:changeToCorrect}.

\begin{figure}[htpb]
	\centering
	\includegraphics[width=1.0\columnwidth]{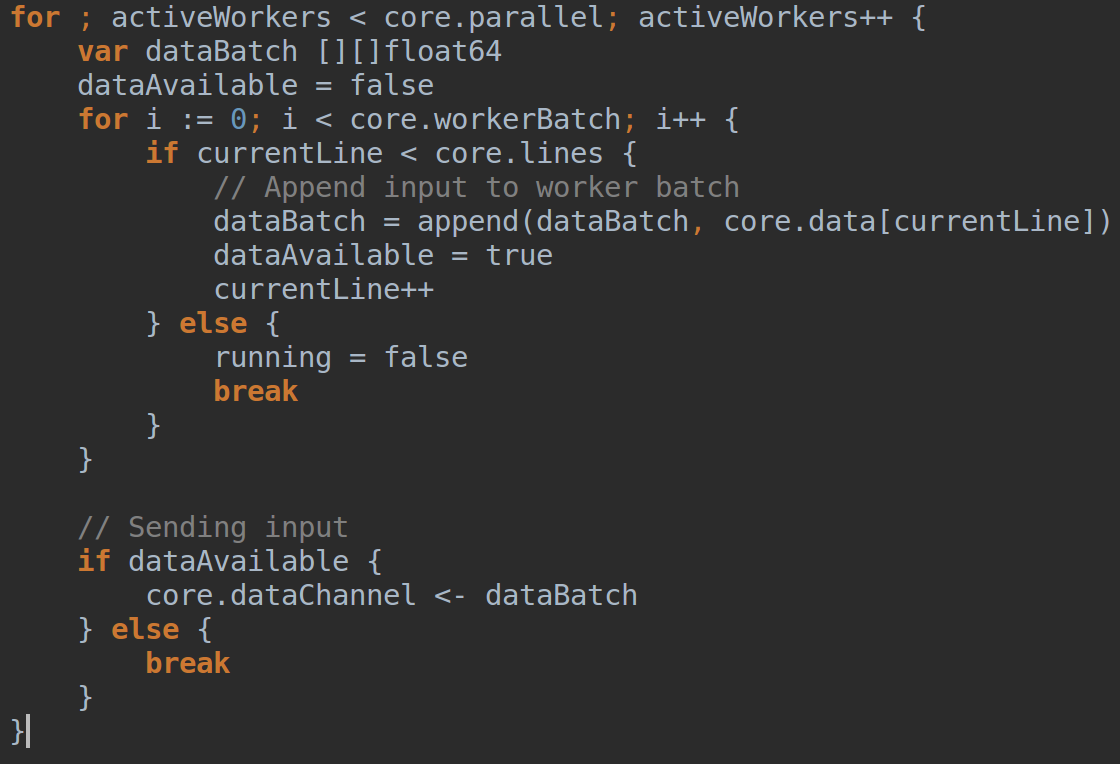}
	\vspace{-1em}
	\caption{Old code, appending to a new slice}
	\vspace{-0.5em}
	\label{fig:oldCode}
\end{figure}

\begin{figure}[htpb]
	\centering
	\includegraphics[width=1.0\columnwidth]{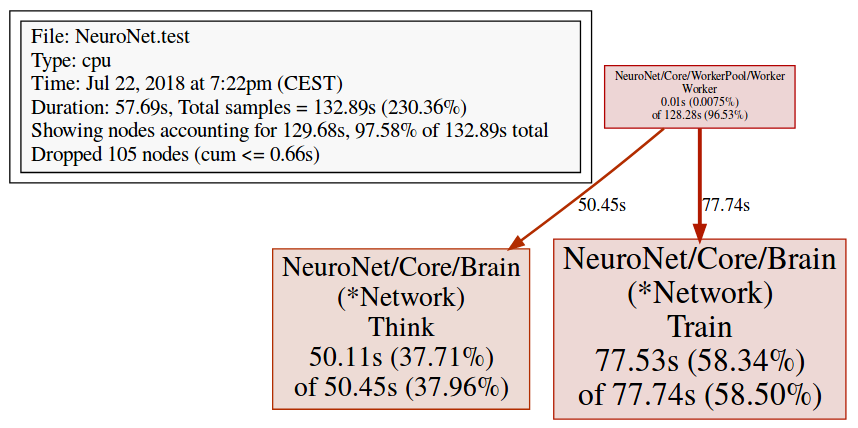}
	\vspace{-1em}
	\caption{CPU profile before code changes}
	\vspace{-0.5em}
	\label{fig:pprofOld}
\end{figure}

In example the code in \autoref{fig:oldCode} takes 129.68 seconds for 1 training and 1 testing with the MNIST dataset, 4 cores, and a worker batch setting of 100, as shown in \autoref{fig:pprofOld}.

\begin{figure}[htpb]
	\centering
	\includegraphics[width=1.0\columnwidth]{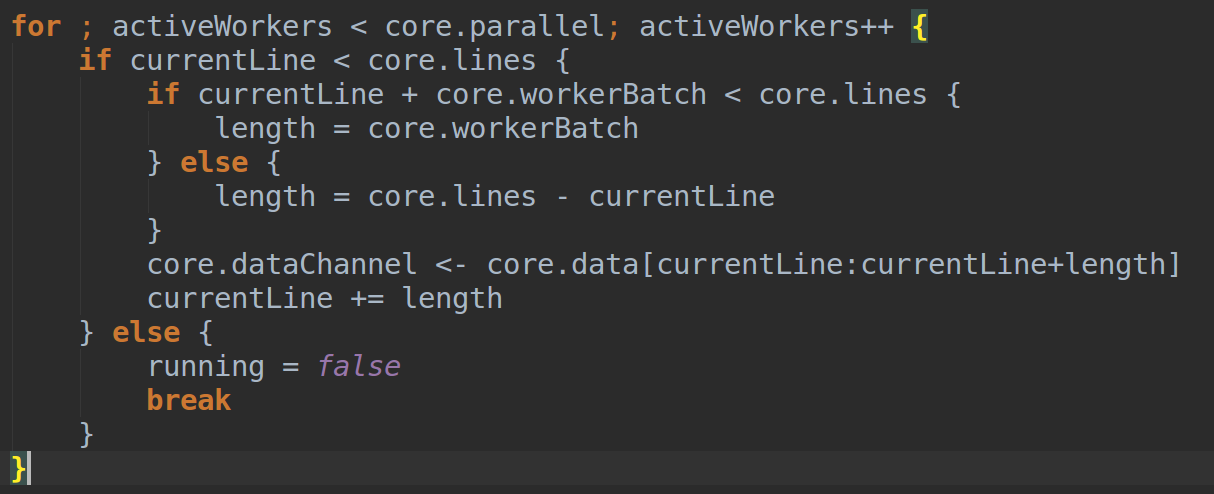}
	\vspace{-1em}
	\caption{New code, using the slice functionality}
	\vspace{-0.5em}
	\label{fig:newCode}
\end{figure}

\begin{figure}[htpb]
	\centering
	\includegraphics[width=1.0\columnwidth]{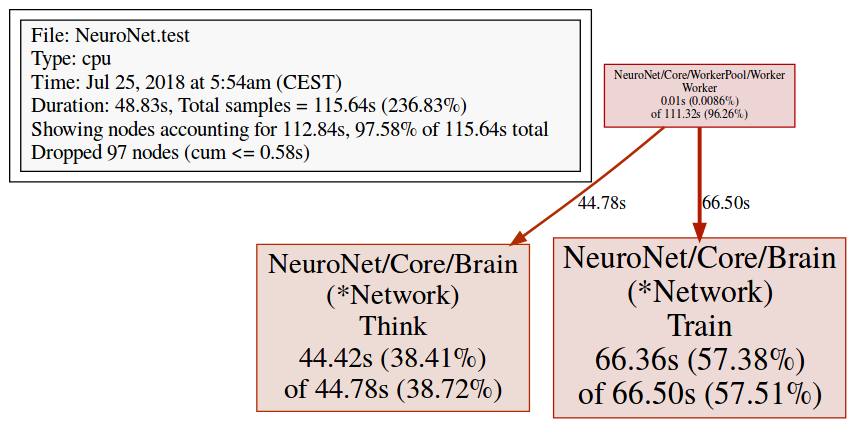}
	\vspace{-1em}
	\caption{CPU profile after code changes}
	\vspace{-0.5em}
	\label{fig:pprofNew}
\end{figure}

In comparison when utilizing slices instead of making a slice and appending the data lines within core.run() to send the worker batches to the workers as shown in \autoref{fig:newCode} saved about 17 seconds on the Lenovo, as is visible in \autoref{fig:pprofNew}.

\begin{figure}[htpb]
	\includegraphics[width=1.0\columnwidth]{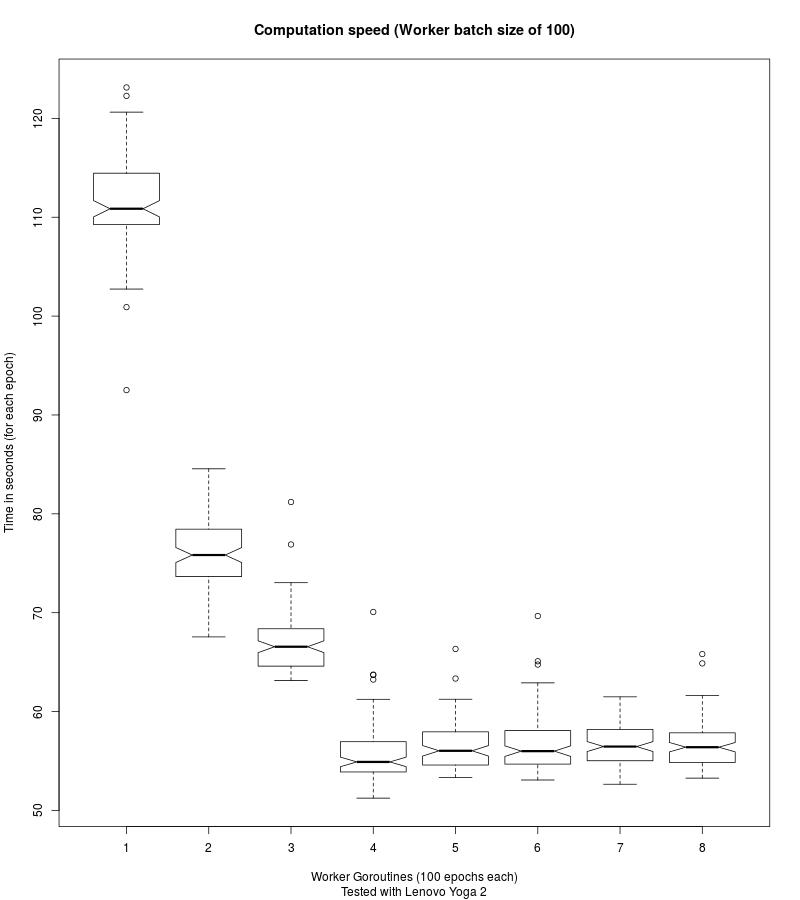}
	\vspace{-1em}
	\caption{Benchmark before code changes}
	\vspace{-0.5em}
	\label{fig:oldAllGoroutines}
\end{figure}

\begin{figure}[htpb]
	\includegraphics[width=1.0\columnwidth]{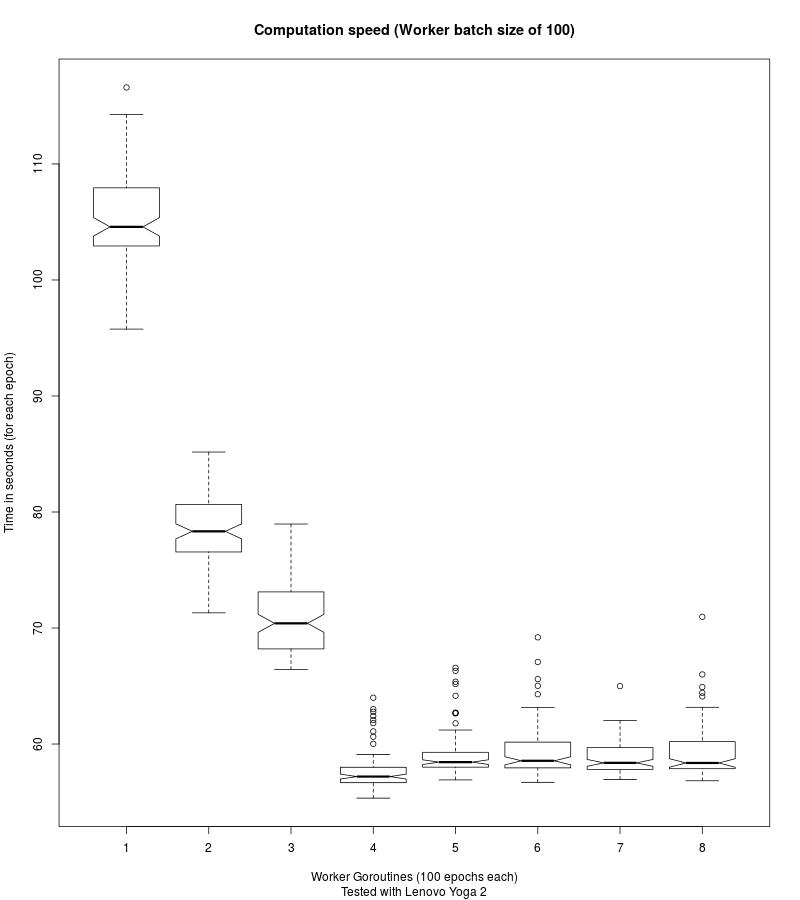}
	\vspace{-1em}
	\caption{Benchmark after code changes}
	\vspace{-0.5em}
	\label{fig:newAllGoroutines}
\end{figure}

But when comparing the old speedup in \autoref{fig:oldAllGoroutines} with the new speedup in \autoref{fig:newAllGoroutines}, the computation speed is on average a little slower. The assumption is that - due to code optimizations and therefore less workload - the CPU chose to use a lower clock speed, so computation took longer. This assumption is untested, but context switching can be ruled out with a high certainty as the possible speedup of the new code has been tested with 4 cores: If context switching would have any negative impact, it should have shown in this test.

\subsubsection{Never trust a single source}
"Why is this important, when the objective is just to write code?" one may ask. The answer is pretty straightforward and was also a pitfall in the early stages of this work.

When implementing software to solve a certain problem, there are often different sources available. It is paramount to not trust a single source. If possible, look at the original studies, try to get the original data - but watch out for personal interests and possible skewed or even fabricated data. Therefore also look for meta analyses and systematic reviews, which look at a broader spectrum, the methods used, and other vital data points to find eventual outliers or problems with the data. Also don't trust sources that are given to you by a single source. Check it first. Everything else costs time and nerves: Only assuming, but not knowing, that the source is correct can lead to triple-checking code, math, and data, even though the error lies somewhere else. Two examples follow.

Many different sources in the web, including Wikipedia, managed to quote the correct source for linear unit activation functions, but used the old equations from the first two versions of the paper. There is a small effect on the derivative functions when x = 0. In example with ELU: The old and new equation are only equal when $\alpha$ = 1. Otherwise, when x = 0, f'(x) should be $\alpha$, not 1.

One trusted source - due to one University module handed out an excerpt without citing the source which was the basis for an assignment to calculate a forward, backward, and forward propagation by hand, and creating a simple neural network, which both got graded - used the updated weights as basis for the backpropagation. 

\begin{figure}[htpb]
	\includegraphics[width=1.0\columnwidth]{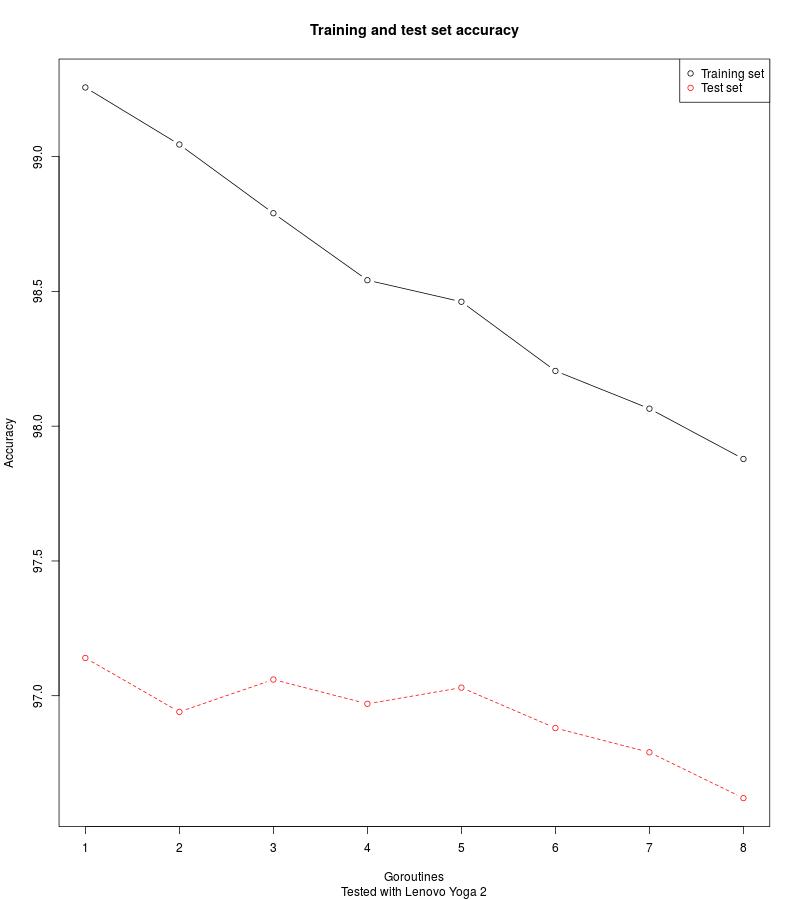}
	\vspace{-1em}
	\caption{Old Plot showing the accuracy decrease}
	\vspace{-0.5em}
	\label{fig:oldPlot}
\end{figure}

There was a larger accuracy drop\autoref{fig:oldPlot} when using multiple goroutines.

\begin{figure}[htpb]
	\includegraphics[width=1.0\columnwidth]{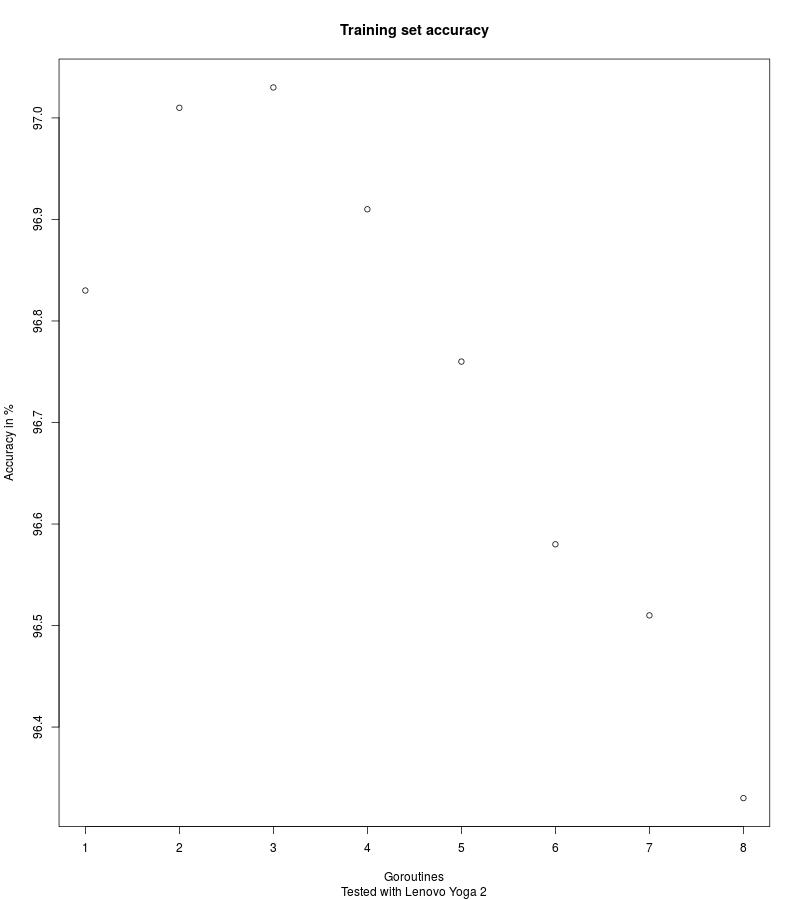}
	\vspace{-1em}
	\caption{Accuracy with wrong parameters}
	\vspace{-0.5em}
	\label{fig:newPlot}
\end{figure}

Although it was possible to see when to tweak the parameters to gain a higher accuracy with a single core\autoref{fig:newPlot}, there has been found no practical use of these values for the correct implementation - the hyperparameters vary widely, so they have to be tuned differently.

Also mathematical sources are often not the best source for calculations in information systems. Math has been spared the problem of both errors due to overflow and underflow (except when using calculators or information systems). Also there is no need to optimize for memory or computation speed.

The problem with trusting the wrong data has been solved with further research from different sources and consulting a mathematician to check if the partial derivatives and all formulas have been implemented correctly. The error has then been found very quick when checking against the standard reference\cite{GoodfellowEtAl2016}.

\subsection{Comment on Go}
Go is a wonderful language to write code. Implementation and testing of the neural network seemed to be easier than with other programming languages. But go also has some drawbacks (as does any language).

The main annoyance were the "unused imports" bugs. Sometimes only certain outputs are needed for testing which will get dropped by the developer immediately afterwards. It's good that the Go compiler sees these oversights as errors, even though they are a huge annoyance. A probably better way would be if unused imports won't be tested in a debug environment, only in production. But this would have additional drawbacks when Go is used in environments where code quality is not highly valued.

Another annoyance is the "sudden 'bad file descriptor' of doom". Sometimes it's just a "data reader error: file already closed". It was not possible to pin down what exactly causes the error, only that it affects the file as a whole. Not even deleting and creating a new file with the same name helps to overcome that error. Further testing is needed.

An additional observation that can ruin ones day is, that the Go compiler for some reason accesses trashed files, at least under Linux. There is no problem when files are overwritten by new files. But if a file gets deleted, and a new one inserted instead, Go sometimes seems to try to compile the deleted files, which can lead to hard to trace errors. If there's, in example, an error where the Go compiler expects an integer value, the code provides an integer value, but recently a file with the same function expecting a double value had been trashed, simply empty the trash bin.

Another "hard to debug except when you know it" part is: "panic: runtime error: invalid memory address or nil pointer dereference". This error occurs when the object has not been created with new(...). If it's further up in the code, i.E. some struct attribute, this error is not easy to find. When starting with Go that panic tells almost nothing about its nature.

Circular dependencies are not allowed. They can happen while refactoring code or when making some design mistake. It's good that Go does not permit them as they are a sign of bad software design.

The short variable declaration := is very handy. Go recognizes the type and assigns the value to the left hand variable. The best part: It won't break type safety, 
which prevents weird behavior.

With the test coverage profiler it is easy to see the current code coverage. There is also the possibility to create test heat maps and to show the test coverage in the browser with highlighting good, poor, or not covered code parts\footnote{Go test coverage and html heatmap: \url{https://blog.golang.org/cover}}.

There are memory and cpu profilers, and even a profiling tool\footnote{Go profiling: \url{https://blog.golang.org/profiling-go-programs}}. It is easy to list the top cpu or memory consumers or show a profiling web. Therefore memory issues can be found easily, as well as slow code parts.

Go uses function inlining which is a great method for speeding up code.

Goroutines are very lightweight\footnote{Currently 2, 4, or 8 kB per Goroutine, depending on the version, i.E. https://github.com/golang/go/issues/7514}. As they're very efficient and only start to run if they get data from a channel, there's the probability of an application for parallelized neurons instead of only parallelized networks.

It's easy to find and fix race conditions with Go as it comes with its own race detector.

\section{Conclusion and Future work}
It has been learned how to use the programming language Go and about its parallel speedup possibilities.
The main accomplishment of this work is to have managed to create a stable and fast neural network.
The hardest part was to understand the mathematical concepts and ramifications behind neural networks and how to implement them software wise.

The main focus of this work was to see how the parallel speedup of a neural network behaves with the language Go. Due to time and resource restrictions only little derivations from the main focus were made. There are still ways left to make this neural network even more efficient, with higher accuracy, and so on. The current version could have some possible memory leaks. They will be fixed in a future version. As there will be further changes due to development and additional insights, the code will probably be refined and refactored in the future.

Some parts of the code are still untested - mainly file reading and writing. As they work as intended no additional effort has been made to get 100\% test coverage in these areas. Here is room for improvement.

Optimization of the neural network would be the largest part of the future work. Currently it is only a simple network with Bias. It would be possible to implement momentum\cite{LeCun2012} and other artifacts to achieve higher accuracies. NADAM and other stochastic gradient descent optimization algorithms\cite{DBLP:journals/corr/Ruder16} could be implemented too.

Smaller changes will also include several options, in example if the user wants bias nodes, which error severity to log, and to choose different lambdas for the L1 and L2 Regularization in the elastic net. Adaptive learning rates\cite{LeCun2012} would be of interest too. Different loss functions, especially Cross Entropy Loss\cite{sadowski2016notes} will be implemented in the future.

There is an interest to look into Self-Normalizing Neural Networks\cite{DBLP:journals/corr/KlambauerUMH17}.




\end{document}